\documentclass[10pt,twocolumn,letterpaper]{article}

\usepackage{wacv}
\usepackage{times}
\usepackage{epsfig}
\usepackage{graphicx}
\usepackage{amsmath}
\usepackage{amssymb}
\usepackage{subfigure}
\usepackage{comment}
\newcommand{\bx}{\mathbf{x}}

\newcommand{\by}{\mathbf{y}}
\newcommand{\bk}{\mathbf{k}}
\newcommand{\bs}{\mathbf{s}}
\newcommand{\cR}{\mathcal{R}}
\newcommand{\bh}{\mathbf{h}}
\usepackage[ruled,vlined]{algorithm2e}

\def\wacvPaperID{1371} 

\wacvfinalcopy 
\ifwacvfinal
\def\assignedStartPage{9876} 
\fi

\ifwacvfinal
\usepackage[breaklinks=true,bookmarks=false]{hyperref}
\else
\usepackage[pagebackref=true,breaklinks=true,colorlinks,bookmarks=false]{hyperref}
\fi

\ifwacvfinal
\else
\fi

\begin{document}

\title{ Improving Robustness and Uncertainty Modelling in Neural Ordinary Differential Equations}

\author{
Srinivas Anumasa\\
Computer Science and Engineering \\
Indian Institute of Technology Hyderbad, India\\
{\tt\small cs16resch11004@iith.ac.in}
\and
P.K. Srijith\\
Computer Science and Engineering \\
Indian Institute of Technology Hyderbad, India\\
{\tt\small srijith@cse.iith.ac.in}
}

\maketitle

\begin{abstract}
Deep learning models such as Resnets have resulted in state-of-the-art accuracy in many computer vision problems. Neural ordinary differential equations (NODE) provides a continuous depth  generalization of Resnets and  overcome drawbacks of Resnet such as model selection and parameter complexity. Though NODE is more robust than Resnet,  we find that NODE  based architectures are still far away from providing robustness and uncertainty handling required for many computer vision problems. We propose novel NODE models which address these drawbacks. In particular, we propose Gaussian processes (GPs) to model the fully connected neural networks in NODE (NODE-GP) to improve robustness and uncertainty handling capabilities of NODE. The proposed  model is flexible to accommodate different NODE architectures, and further improves the model selection capabilities in NODEs. We also find that numerical techniques play an important role in modelling NODE robustness, and propose to  use different numerical techniques to improve NODE robustness. We demonstrate the superior robustness and uncertainty handling capabilities of proposed models on adversarial attacks and  out-of-distribution experiments for the image classification tasks. 

\end{abstract}

\section{Introduction}

The ability of deep learning models to capture rich representations of high dimensional data has lead to successful application   in computer vision problems like image classification~\cite{huang2017densely,he2016resnet}, image captioning~\cite{Vinyals_Toshev_Bengio_Erhan_2015}. The backbone of many of the recent computer vision tasks are deep learning mode1s such as Resnets~\cite{he2016resnet}. They allowed deep learning models to solve complex computer vision tasks by training deep neural networks with more than 100 layers without suffering from vanishing gradient problem. Resnets achieve this using residual connections where input at any layer is added to the output of that layer.  

Recently, generalization of Resnet based models was introduced,  inspired by ordinary differential equations (ODE)~\cite{node,anode}. An ODE parameterized  by a neural network can be seen to generalize the Resnets to arbitrarily or infinitely many layers. Such neural ODE (NODE) models have been shown to achieve performance close to Resnet but with a smaller number of parameters. In addition, model selection is also easy with these models as the numerical solver for ODE can automatically determine the number of layers. Like in Resnets, the representations learnt through NODE block are finally mapped to the output through a fully connected neural network (FCNN).  NODEs are also shown to be more robust than convolutional neural networks (CNN)~\cite{robustode}. 

To achieve a generalization performance similar to Resnet architecture, variants of NODE were proposed~\cite{anode,augmentedode}. They improve generalization performance by concatenating ODE blocks or by adding extra dimensions.  When these models are used  in high risk domains such as  medical~\cite{cell,tumour} and autonomous driving~\cite{fridman2019autonomous},  the  model should be robust and be able to handle uncertainty well. However, we find that the  neural ODE model  fails to achieve robustness and  uncertainty modelling capabilities required for these real world applications. They perform poorly on adversarial attacks which injects adversarial noise to the data~\cite{FGSM,pgd}. They may classify an out-of-sample data to a wrong class with high probability.  We propose to address these drawbacks in Neural ODE through the use of Bayesian non-parametric approaches like Gaussian processes and through more stable numerical solvers for NODEs.

Bayesian  models exhibit good robustness and excellent uncertainty modelling capabilities~\cite{gelman04,gal2016uncertainty}. In particular, Gaussian processes (GPs)~\cite{williams2006gaussian} are useful for modelling uncertainty due to their fully  Bayesian non-parametric nature.  GPs can accurately model predictive uncertainty and provide a more meaningful predictive probabilities. 
Further, it reduces model selection efforts to a great extent and facilitates hyper-parameter estimation  through the marginal likelihood.  These properties have lead to their use in various real world problems~\cite{Rodrigues_Pereira_2018}, active learning~\cite{Kapoor_Grauman_Urtasun_Darrell_2009}  and global optimization such as Bayesian optimization~\cite{Shahriari_Swersky_Wang_Adams_de_Freitas_2016}.  

We propose a flexible and scalable approach to incorporate uncertainty modeling capabilities in NODE by replacing the final fully connected neural networks with GPs (NODE-GPs). Alternatively, this can be seen as a GP whose kernel takes feature representation provided by NODE as input. Such an approach is shown to have very good robustness and uncertainty modeling capabilities without affecting generalization performance of deep learning models~\cite{DKL,SVDKL}. Moreover, this will also contribute towards  reducing model selection efforts by avoiding the requirement to select the FCNN architecture.
Thus, the proposed approach further elevates the advantages of NODE for deep learning through better uncertainty modelling and reduced model selection effort. Further, the proposed model offers flexibility to incorporate different kinds of NODE architectures such as ANODE~\cite{anode} and numerical techniques easily. 

To the best of our knowledge, the  role  of  numerical methods on robustness capabilities of NODE have not been studied in the literature before. We investigate different numerical techniques and show that the robustness properties of the NODE are affected by the numerical method used. We demonstrate that NODE using higher order numerical methods are more robust.
To trade-off robustness and computational cost, we propose a mixed order numerical technique for NODE  with different NODE blocks using different order numerical methods. The proposed mixed order numerical technique for NODE which when combined with GPs showed a superior robustness and uncertainty modeling capabilities
 on adversarial attack and out-of-distribution (OOD) experiments on image classification problems.

Our contributions can be summarized as follows:
\begin{itemize}
    \item {Propose a novel model which combines NODE based architecture and GP (NODE-GP) to improve  uncertainty handling  in NODE.} 
    \item {Demonstrating that  numerical method  affects NODE robustness and proposing a mixed order numerical method for NODE to improve its robustness.  }
    \item {Using  mixed order numerical technique in   NODE-GPs and  demonstrating the superior  robustness  and  uncertainty modelling capabilities through adversarial attacks and OOD experiments.}
\end{itemize}
In the following section, we introduce  the necessary background required, followed by proposed methodology and experimental setup. In the experiments section, we demonstrate the robustness and uncertainty handling  capabilities of the proposed model on image classification.  

\section{Related Work}
NODE~\cite{node} is a generalization of Resnet~\cite{he2016resnet} as continuous approximation of intermediate feature vectors of a Residual block. In~\cite{node} a memory efficient adjoint method was proposed for computing the gradients of parameters.  This  approach lead to a series of works~\cite{anode,augmentedode,neuralspline} by addressing the issues in ~\cite{node}. Stochastic variants of NODE~\cite{neuraljump,stabilizingneural} were also proposed but the evaluation of their proposed methods are restricted to simple architectures. To achieve similar generalization performance compared with Resnet architecture, \cite{anode} proposed a modified adjoint approach which address the issue of computing incorrect gradients in~\cite{node}. The  approach~\cite{anode} allow concatenation of NODE blocks to achieve a similar accuracy compared to an equivalent Resnet architecture, but is restricted to the usage of discrete numerical methods with fixed step-sizes. It was shown in~\cite{robustode} that NODE~\cite{node} is robust to adversarial attacks compared to similar size DNN architecture, but again restricted to simple architecture. Although~\cite{anode} achieves a good generalization performance, we show in our experiments  ~\cite{anode} can be easily fooled with adversarial attacks and lack  uncertainty modeling capabilities. 
 \section{Background}
Let $\mathcal{D} =  \{X,\by\} = \{(\mathbf{x_i},y_i)\}_{i=1}^{N}$ be set of training data points with $\mathbf{x_i} \in \cR^D$ and $y_i \in \{1,\ldots,C\}$. We denote the test data point as $(\bx_*, y_*)$. The aim is to learn a function which maps input from the data $\mathbf{x}$ to a class label $y$  so that it will have good generalization performance.  Let $g$ be the function learnt using a neural network model, while the function learnt using Gaussian processes to be denoted by $f$. For a deep learning model such as Resnet composed of multiple  blocks, we denote a  block $i$  as $g_i (\bx,\pmb{\theta }_i)$. The hidden layers in a neural network are denoted as $\mathbf{h}_i$. 
In this section, we will provide background information required to understand the  proposed model. 
\subsection{Ordinary Differential Equation(ODE)}
Ordinary differential equations (ODE) occur in many real world problems like predicting the motion of objects, and change in rate of GDP. For instance, an ODE is written as 
\begin{equation}
\frac{d\bs_t}{dt} =  g(t,\bs_t)    
\end{equation}
where $\bs_t \in \mathbb{R}^d$,  $g :\mathbb{R}^{d+1}\to  \mathbb{R}^d$. 
We will  focus on initial value problems (IVP), i.e. given initial state $\bs_0$ we want to compute state value $\bs_T$ at time $T$.
\begin{equation}
    \bs_T = \bs_0 + \int_{0}^{T}g(t,\bs_t)dt
\end{equation}
Unfortunately, solutions for most of the ODE cannot be obtained in a closed form. Numerical methods such as the Euler method and Runge-Kutta4 (RK4) come for the rescue to approximate the solution. 

\begin{figure}
\centering 
\includegraphics[scale=0.3]{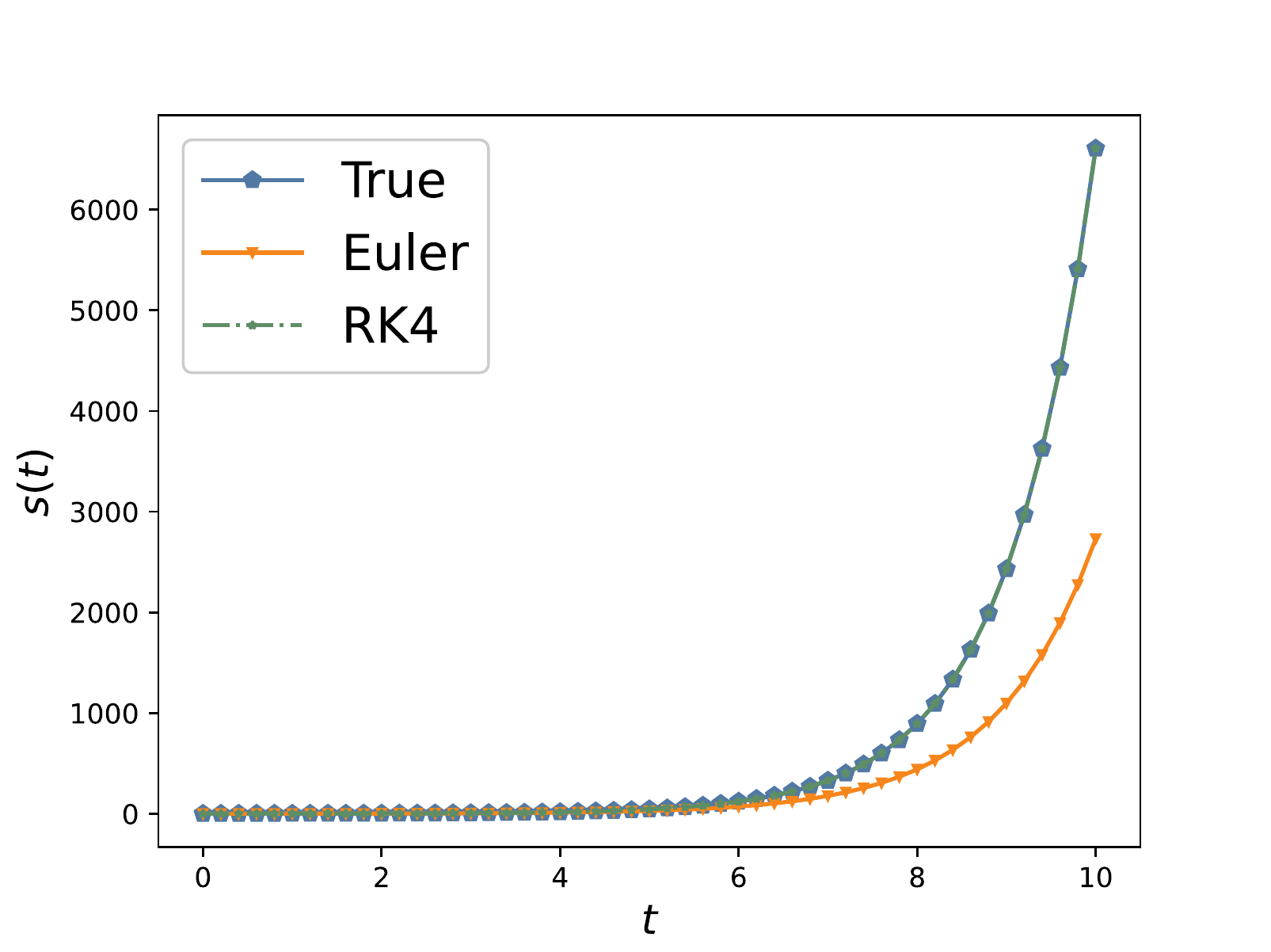}
\caption{Comparing the solutions computed using Euler and RK4 }
\label{numerical}
\end{figure}

The Euler method is a simple one step method. The parameter required is the step-size $dt$. The final value $\bs_T$ is obtained by iteratively updating  the values as
\begin{equation}
\bs_{t+1} = \bs_t + dt \; g(t,\bs_t)
\end{equation}
RK4 method is a four step method and approximates the solution closer to the actual compared to Euler method.
\begin{eqnarray}
& \bk_1 = g(t,\bs_t) \quad  ; \quad 
\bk _2 = g(t+\frac{dt}{2},\bs_t + dt\frac{\bk _1}{2}) \nonumber \\
& \bk _3 = g(t+\frac{dt}{2}  
\bs_t + dt\frac{\bk _2}{2}) ;
\bk _4 = g(t+dt,\bs_t+dt\bk_3) \nonumber \\
& \bs_{t+dt} = \bs_t + dt\frac{1}{6}(\bk_1 + 2\bk_2 + 2\bk_3 +\bk_4)
\end{eqnarray}
Here, $\bk_1$ is the slope at point $(t,\bs_t)$, $\bk_2$ is an estimate of the slope at the midpoint $(t+\frac{dt}{2}, \bs_t + dt\frac{\bk _1}{2})$ obtained  using slope $\bk_1$ to step half-way through the time step. Similarly, $\bk_3$ and $\bk_4$ are obtained 
and the next value $s_{t+dt}$ is obtained by taking a step towards the weighted average of slopes.
To illustrate the differences between RK4 and Euler numerical methods, let us consider a simple ODE $\frac{ds}{dt} = \bs_t$ with given initial value at $t=0$ as $\bs_0$. True solution of the given ODE is $\bs_t = \exp(t)\bs_0$.  Figure ~\ref{numerical}  illustrates the solutions computed using RK4 and Euler method,  we can observe RK4 method is overlapping with the true solution whereas solution using  Euler method is moving further away from true solution over time.

\subsection{Neural Ordinary Differential Equations}
Deep learning models such as Resnets learn a sequence of  transformation by mapping input $\mathbf{x_i}$ to output $y_i$. In a Resnet block, computation of a hidden layer representation  can be expressed using the following transformation. 
\begin{equation}
    \textbf{h}_{t+1} = \textbf{h}_t + g_t(\textbf{h}_t,\pmb{\theta}_t)
    \label{eqn:resnet}
\end{equation}
where $\textbf{h}_t$ is a feature vector with $t\in\{0...T\}$ and $g$ is a neural network parameterized by parameters $\pmb{\theta}_t$. If we use the same transformation at every step, Equation \ref{eqn:resnet} can be written as 
\begin{equation}
    \textbf{h}_{t+1} = \textbf{h}_t + g(\textbf{h}_t,\pmb{\theta}),
    \label{Eulerstep}
\end{equation}
and this is equivalent to computing the trajectory of the following ODE using Euler method with step size one. 
\begin{equation}
    \frac{d\textbf{h}_{t}}{dt} =  g(\textbf{h}_t,\pmb{\theta})
    \label{Eulerstep1}
\end{equation}

NODE~\cite{node} has taken a step further by modeling continuous transformation of feature vectors in the limiting case of step size tending to zero. Given initial feature vector $\textbf{h}_0$, the final feature vector $\textbf{h}_T$ can be computed by solving the ODE $\textbf{h}'_t = g(\textbf{h}_t,\pmb{\theta})$ parameterized by $\pmb{\theta}$.  On the obtained feature vector $\textbf{h}_T$ necessary transformations are applied using a fully connected neural network (FCNN), involving multiple linear mapping and activations to predict class probabilities. Using cross-entropy loss function and stochastic mini-batch gradients the parameters of the model are updated using memory efficient adjoint based approach. 

Variants of NODE~\cite{anode,augmentedode} are proposed which can achieve a  generalization capability same as Resnets~\cite{he2016resnet}. It consists of multiple NODE blocks, each NODE block $B_i$ given initial feature vector $\textbf{h}_0^i$, using an assigned numerical method computes  intermediate  feature vectors $\textbf{h}_t^i, t \in \{0,1,..T\}$  using the function $g_i(\textbf{h}_t^i,\pmb{\theta}_i)$ parameterized by $\pmb{\theta}_i$. The feature vector $\textbf{h}_T^i$  obtained from current block $B_i$ then undergo  necessary transformation to match the dimension required for processing in block $B_{i+1}$. The final feature vector  from the final Block is transformed through a FCNN to predict class probabilities. Still, NODE and its variants lack the  uncertainty modelling  and robustness capabilities as observed in our experiments.   

\subsection{Gaussian process}
Gaussian processes (GPs) provide a Bayesian non-parametric approach to learning functions. A GP prior defined over the function can determine various properties of the functions like their smoothness, and stationarity. This can be specified using the mean function $m(\bx)$ and  covariance function or kernel $k(\bx,\bx')$ associated with a GP.  A commonly used kernel is the radial basis function (RBF) kernel $k(\bx,\bx') = \exp(-\frac{1}{2 \ell^2} ||\bx - \bx'||^2)$, where $\ell^2$ (length) represents the wiggliness of the function and is learnt from the data.  Assuming  $f(\bx)$ is a function sampled from a zero mean GP, we can write $\mathbf{f} = f(X)$ to follow a Gaussian distribution.  
$\mathbf{f} \sim \mathcal{N}(0,k(X ,X))$

In regression, the output $y$ lies around a latent function $f(\bx)$ sampled from a GP as $y|f(\bx) \sim \mathcal{N}(y;f(\bx),\sigma^2\it{I})$.
The posterior distribution over the latent function is obtained by combining the prior over $\bf{f}$ and the likelihood through Bayes theorem. 
 \begin{equation}
 p({\bf f} | {\bf y})=\frac{\prod_{n=1}^{N} p\left(y_{n} | f(\bx_{n})\right) p(\bf{f})}{p(\bf{y})}    
 \end{equation}

The learning in GPs involves estimating hyper-parameters like $\ell^2$ from the marginal likelihood, $p(\by) \sim \mathcal{N}(0, k(X,X) + \sigma^2 I)$. This involves a complexity of $O(N^3)$ due to matrix inversion  $(K_{X,X} + {\sigma^2 I})^{-1}$ over $N\times N$ matrix, where $N$ is the number of training samples. To avoid the cubic complexity sparse approximations of full GP were proposed which  reduces the complexity to $O(M^2N)$, where $M<<N$ is the number of inducing inputs~\cite{snelson2006sparse}.
For the classification task, the likelihood of the model $p(\by |f(\bx))$  is a softmax function and models the probability of generating the output given the latent function value. 
However, due to the non-Gaussian nature of the likelihood, approximate inference techniques such as variational inference are employed to get the marginal likelihood and posterior~\cite{hensman13,blei2017variational}. 

\subsection{Adversarial Attacks and Uncertainty modelling}
\label{sect:adversarial}
In order to study the robustness of models we consider two scenarios : Adversarial attacks and uncertainty modelling on out-of-distribution data. Adversarial attacks aim to generate examples which can fool deep learning models to predict a wrong class.  We discuss two adversarial attacks  namely Fast Gradient Sign Method (FGSM)~\cite{FGSM} and  Projective Gradient Descent (PGD)~\cite{pgd} which are helpful to determine robustness of the model.
\begin{itemize}

\item \textbf{Fast Gradient Sign Method(FGSM)} It is a one step  white-box adversarial attack, where the adversary has access to the entire model. Given a sample image $\bx_*$, gradient of loss  with respect to each pixel $\nabla_{\bx_*} L (\pmb{\theta},\bx_*,y_*)$ is computed and a proportion of the computed gradient is added to the original image. Thus, $\Tilde{\bx}_* = \bx_* + \epsilon \; sign(\nabla_{\bx_*} L (\pmb{\theta},\bx_*,y_*))$.  It involves a hyperparameter  $\epsilon$  which is the strength of the gradient added to the image~\footnote{As our data set are normalized, we test the model performance on $\epsilon$ ranging from $0.05$ to $0.3$.}. 
\item \textbf{Projective Gradient Descent(PGD)} Unlike FGSM attack PGD attack is considered to be a powerful white-box attack. PGD is an iterative algorithm, at every step $\bx_*^{t+1}$ it computes the gradients of loss with respect to each pixel $\nabla_{\bx_*} L (\pmb{\theta},\bx_*,y_*)$, and a proportion of the computed gradient is added to the original image to get an adversarial image. To avoid too much dissimilarity between the actual and perturbed, the perturbed image is projected back into a feasible set around the actual image based on the provided $l_p$ norm. In our experiments we choose $l_\infty$ norm. We have to provide necessary hyperparameters like number of steps $n$,radius $\epsilon$ used for constructing the feasible set  $S(\epsilon)$  around the sample $\bx_*$ with  induced  norm which we choose $l_\infty$~\footnote{In our experiments, we choose $n=10$,  $\epsilon$ value varies from $0.05$ to $0.3$ and the step size $\eta$  is $0.01$.}. Assuming $\prod$ as projection operator, $\bx_*^{t+1} = \prod_{(\bx_*+S(\epsilon))}^{}(\bx_*^t + \eta \; sign(\nabla_{\bx_*} L (\pmb{\theta},\bx_*,y_*)))$.

\end{itemize}

\subsubsection{Uncertainty Modelling}
DNN models excel in generalization when the testing sample $(\bx_*,y_*)$ are from the same as training data distribution. An ideal model should exhibit the necessary feedback when models are fed with test samples from a different distribution. We want our model to exhibit high uncertainty in this case. We discuss two metrics to measure uncertainty,  Variation Ratio and Entropy. 
\begin{itemize}
    \item \textbf{Variation Ratio(VR):} It is a measure of spread around the mode of class probabilities. Variation ratio can be computed as follows
    $VR[\bx_*] := 1-\max_{y_*}p(y_*|\bx_*)$. 
    If the datapoint is not from the training data distribution or model is uncertain about its true class, we expect to have lower predictive probability, and consequently a larger VR score.
    \item \textbf{Entropy:} It captures the information content and uncertainties in the predictive distribution.  It measures the spread of the probability distribution not just over the maximum class but over all the classes. Entropy is computed as
   $ H[\bx_*] := -\sum_{c}p(y_{*c}|\bx_*)\log p(y_{*c}|\bx_*))$ and is maximum when the predictive distribution is uniformly distributed across the classes and close to zero when it is high for one particular class.
\end{itemize}
\begin{figure*}
\begin{center}
\includegraphics[scale=0.4]{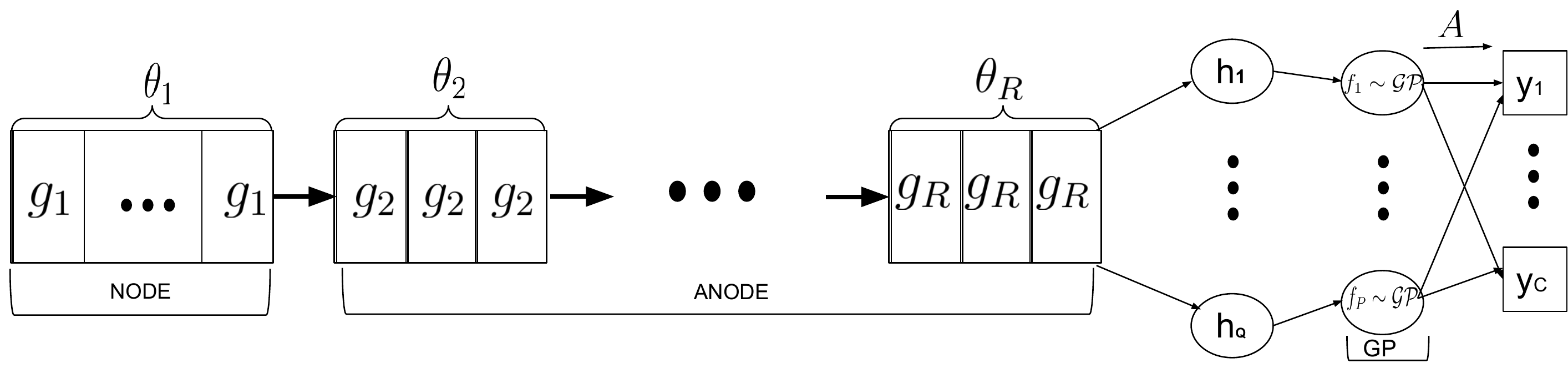}
\end{center}
   \caption{Proposed NODE-GP model combining NODE network with GPs in the final layer.}
   \label{model}
\label{fig:short}
\end{figure*}

\section{Proposed Methodology}

In this section we propose a  novel model  which inherits the properties of NODE based deep learning models and Bayesian non-parametric Gaussian process. This model addresses the drawbacks of NODE and equivalent Resnet in terms of uncertainty handling and robustness. We also discuss the impact of numerical techniques used in NODE models for robustness and propose different NODE architectures based on this observation. 


\subsection{Combining Neural ODE with GPs}



Bayesian learning models compute the predictions by taking expectation over posterior distribution~\cite{gelman04}. This allows them to capture the uncertainty in the model prediction and make them more robust against adversarial attacks. In particular, we consider Bayesian non-parametric models such as Gaussian processes, and an approach to combine the predictive modelling capabilities of GPs into Neural ODEs. Combining neural network representations with GPs were shown to have good uncertainty modelling capabilities~\cite{SVDKL}.  

The proposed model NODE-GP replaces the fully connected neural network part in the NODE architecture with Gaussian processes. The neural ODE blocks transforms the input to some latent representation. The latent representation is then passed through the GP layer to predict the output label. NODE-GPs are quite flexible and can consider any variants of the Neural ODE model before passing it through the final GP layer.  Due to the GP final layer, training in NODE-GP is not straightforward. It involves considerable changes in the objective function especially due to the intractable inference in GPs when used for image classification. The loss function considers the GP marginal likelihood or an approximation to it obtained using an appropriate inference technique. The Bayesian non-parametric properties of GPs will also improve model selection capability of NODE and reduce the effective number of parameters.
Figure~\ref{model} provides an outline of our proposed model where  NODE blocks of different complexity are concatenated together with GPs in the final layer to form NODE-GPs. 

From the point of view of GPs, the functions sampled from GPs are now defined on a feature vector obtained through multiple NODE block transformations. 
Let us assume there are $R$ NODE blocks, each with a neural network transformation $g_i(\textbf{h}_t^i,\pmb{\theta}_i)$,  finally resulting   in a  $Q$ dimensional feature vector $\bh_T^R$. Let the final representation over all the data points be represented as $H_T^R$.  Typically,  the size of the representation is high and Gaussian processes struggle with a high dimensional data.  Following \cite{SVDKL},  $P$ independent GPs are considered with  kernels $k^1,k^2.. k^P$ applied on the subsets of features in the $Q$ dimensional feature vector. 
These GP outputs are then linearly mixed using a matrix $A$ of size $P \times C$ and this captures the  correlations across latent function values associated with class labels. The class probabilities are obtained by applying a Softmax function as the likelihood. For a data point, considering one hot encoding of class label as $\by_i \in \{0,1\}^C$, function values from $P$ GPs as $\bf{f}_i$, the likelihood is given by
\begin{equation}
    p(\by_i|\bf{f}_i) = \frac{exp((A\bf{f}_i)^\top\by_i)}{\sum_c exp((A\bf{f}_i)^\top \bf{e}_c}
    \label{eqn:lik}
\end{equation}
where $\bf{e}_c$ is an indicator vector with a value of 1 in dimension $c$ and $0$ elsewhere.

We use sparse GPs to avoid the $O(N^3)$ complexity of the marginal likelihood. It assumes $j^{th}$ GP  with function values over data points $\bf{f}^j$ ($\textbf{f}^{j} =\{f_{ij}\}_{i=1}^n$) are associated with $m$ inducing points $\textbf{u}^j$ and  inducing inputs $Z$. The prior can be represented using inducing points as
\begin{eqnarray}
\hspace{-5mm}& & p(\textbf{f}^j|  \textbf{u}^j) = \mathcal{N}(\textbf{f}^j|k^j(H_T^R,Z)k^j(Z,Z)^{-1}\textbf{u}^j,\Tilde{K}^j),  \\
\hspace{-5mm}& & \Tilde{K}^j=k^j(H_T^R,H_T^R)-k^j(H_T^R,Z) k^j(Z,Z)^{-1}k^j(Z,H_T^R) \nonumber
\end{eqnarray}

As the likelihood (\ref{eqn:lik}) is non-Gaussian the posterior cannot be computed in closed form. To address this, variational inference is considered  which also allows stochastic gradient training~\cite{SVDKL}. It assumes the variational posterior over inducing points are factorized over independent GPs as $q(\textbf{u}) = \prod_j N(\textbf{u}^j|\pmb{\mu}^j,\textbf{S}^j)$,
where $\pmb{\mu}^j$ are variational mean and $\textbf{S}^j$ are  covariance matrix parameters to be learnt. The corresponding variational lower bound which forms the objective function for training NODE-GPs can be written as
\begin{equation}
\label{lower_bound}
\log p(\textbf{y}) \geq E_{q(\textbf{u})p(\textbf{f}|\textbf{u})}[\log p(\textbf{y}|\textbf{f})] - KL[q(\textbf{u}||p(\textbf{u})].    
\end{equation}
The likelihood can be factorized over data instances as $\log p(\textbf{y}|\textbf{f})= \prod_{i=1}^{n}[\log p(\textbf{y}_i|\textbf{f}_i)]$ which helps in stochastic computation of gradients.    The parameters of both NODE model and GP can be learnt jointly by maximizing the lower bound using noisy approximation of gradient on mini batches of data.  
Equation \ref{lower_bound} is intractable due to the expectation term and is approximated using sampling~\cite{SVDKL}. 
Predictions are made using a set of inducing point values and consequently  latent functions values are sampled from the approximated posterior instead of a point estimate which results in better uncertainty modeling and robustness. 

\begin{table*}
    \centering
   \begin{center}
   {\small
    \begin{tabular}{ |p{3.25cm}|p{1cm}|p{1cm}|p{1cm}|p{1cm}|p{1cm}|p{1cm}|p{1cm}| }
 
 \hline
 \textbf{Epsilon} & \textbf{0.05} &\textbf{0.1}&\textbf{0.15}&\textbf{0.2}&\textbf{0.25}&\textbf{0.3}\\
 \hline
 NODE\_NT2 (RK4) & 60.2 & 54.4 & 50.05 & 45.15 & 40.8 & 36.2 \\
 \hline
 NODE\_NT2 (Euler) & 46.8 & 35.7 & 31.4 & 27.7 &24.05 & 19.95 \\
 \hline
\end{tabular} }
 \end{center}
  \caption{Performance of NODE models under FGSM adversarial attack on Cifar10 data. Both the models trained with number of steps 2, Row-1 using Runge-kutta method of order 4 and Row-2 using Euler method }
    \label{tab:FGSM1}
\end{table*}
\begin{figure*}
\begin{center}     
\subfigure[Euler Method]{\label{fig:a}\includegraphics[scale=0.35]{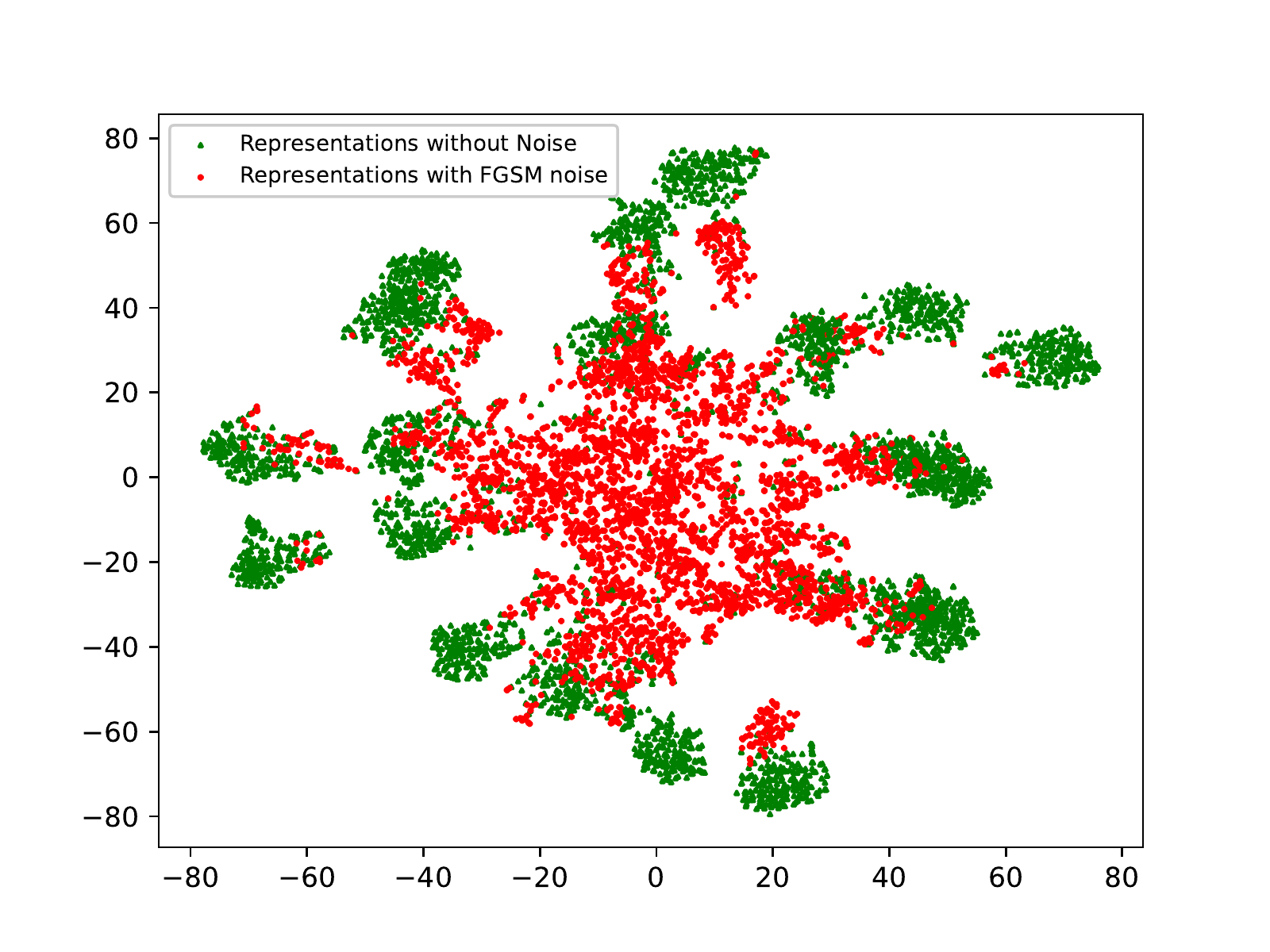}}
\subfigure[RK4 Method]{\label{fig:b}\includegraphics[scale=0.35]{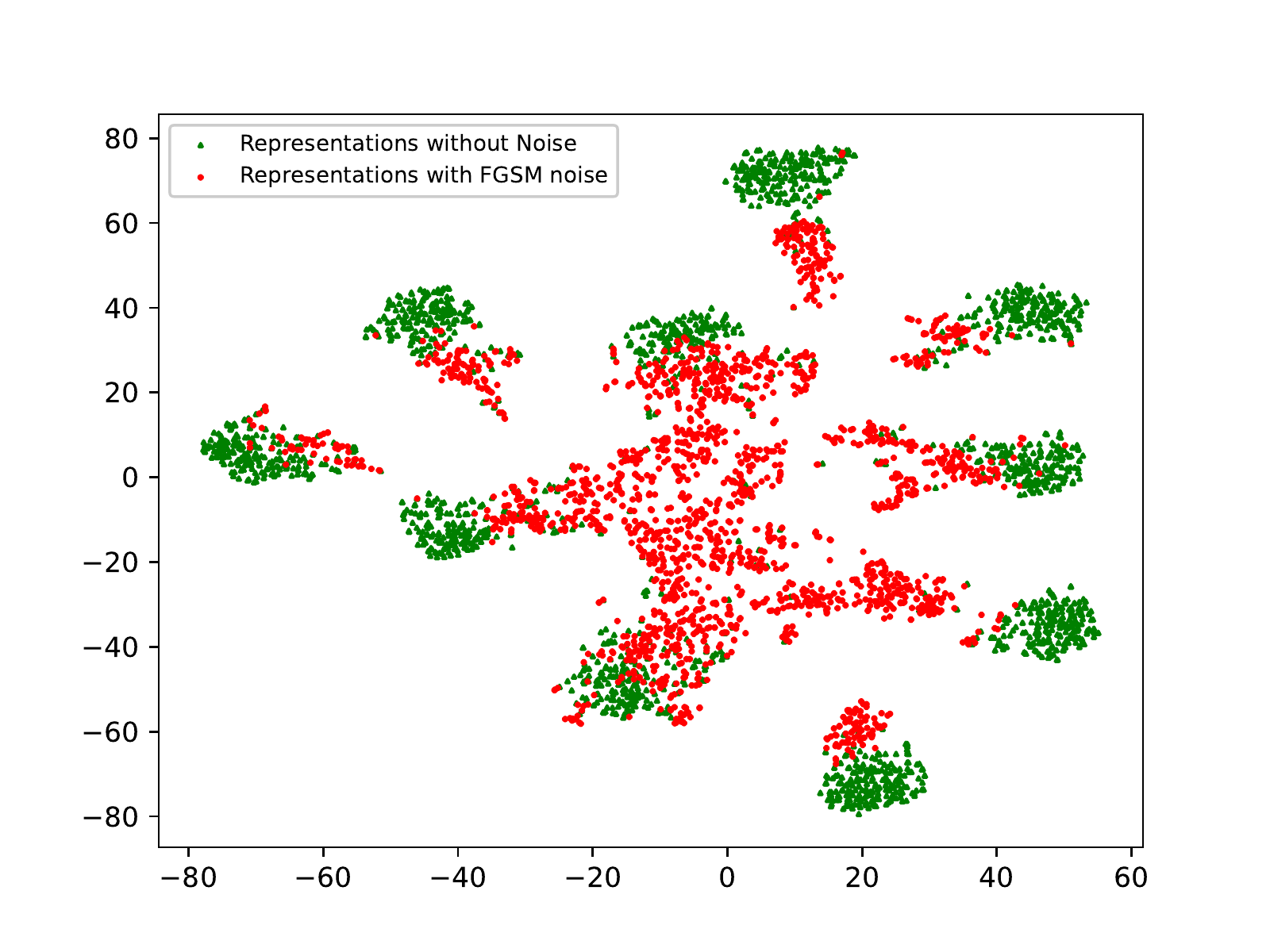}}
\end{center}
\caption{t-sne plots of the features, green points represent feature vectors of clean images and red points represent feature vectors of adversarially affected images (a) ANODE trained with Euler numerical method (b) ANODE trained with RK4 numerical method }
\label{Matters}
\end{figure*}

\subsection{Numerical Methods and Robustness of NODE}
It's found that in ANODE, the choice of numerical method doesn't have much significance in model accuracy~\cite{anode}. First order methods like Euler or Fourth order like RK4 have the same accuracy, but we observe the models vary in their robustness depending on the numerical method used. In~\cite{robustode},  it is shown that NODE~\cite{node} is  robust towards adversarial attacks compared to CNN of similar architecture. In this NODE model, the intermediate feature vectors are computed using an adaptive numerical method~\cite{kimura2009dormand} using a higher order numerical method (RK4/RK5) with step size dynamically changing during the forward propagation. 
In this section, we study and investigate how the order of the numerical technique affects the robustness modelling capabilities in NODE.  

To study robustness,  in Table~\ref{tab:FGSM1} we provide performance of NODE models trained with different numerical methods under FGSM adversarial attack~\cite{FGSM} on the  Cifar10 dataset with increasing \textit{epsilon} values. It can be seen that the performance of ANODE model trained with Euler numerical method degrades more compared to the model trained with RK4 numerical method. 
To determine the variation in performances of Euler and RK4 based NODEs,  we plotted the t-SNE plots  of the representations learned by the models in Figure~\ref{Matters}. In Figure \ref{Matters}(b), we can see the representations(without noise) learned by the model with RK4 numerical method got clustered into 10 clusters which is equal to the number of classes in the Cifar10 dataset. But, this is not the same for NODE models trained using the Euler method. In Figure \ref{Matters}(a), we do not see a clear 10 clusters as compared to RK4 method. This problem in  representation learning without clear distinct boundaries between the classes possibly has led the adversary to easily fool the model. From a numerical method point of view, the solution computed by the Euler method is not close to the actual solution as compared to RK4. Therefore, the path traced by Euler method can be easily altered with a small change in input. 

\subsubsection{Numerical Methods in NODE-GP}
Based on our observation in the previous section, we find that the numerical method used in a NODE plays a role in defining the robustness. But, increasing the order of the numerical method will have an adverse effect on the  computational cost. One should trade-off between computational cost and achievable robustness. We consider different variants of the NODE-GP model differing in the numerical method used in the different blocks of the NODE-GP model. Combining the robustness and uncertainty modelling capability of NODE-GP with an appropriate numerical technique will lead to even more robust models as we observe in the experimental section. 

\begin{figure*}[t]
\centering     
\subfigure[PGD Attack]{\label{fig:a1}\includegraphics[scale=0.38]{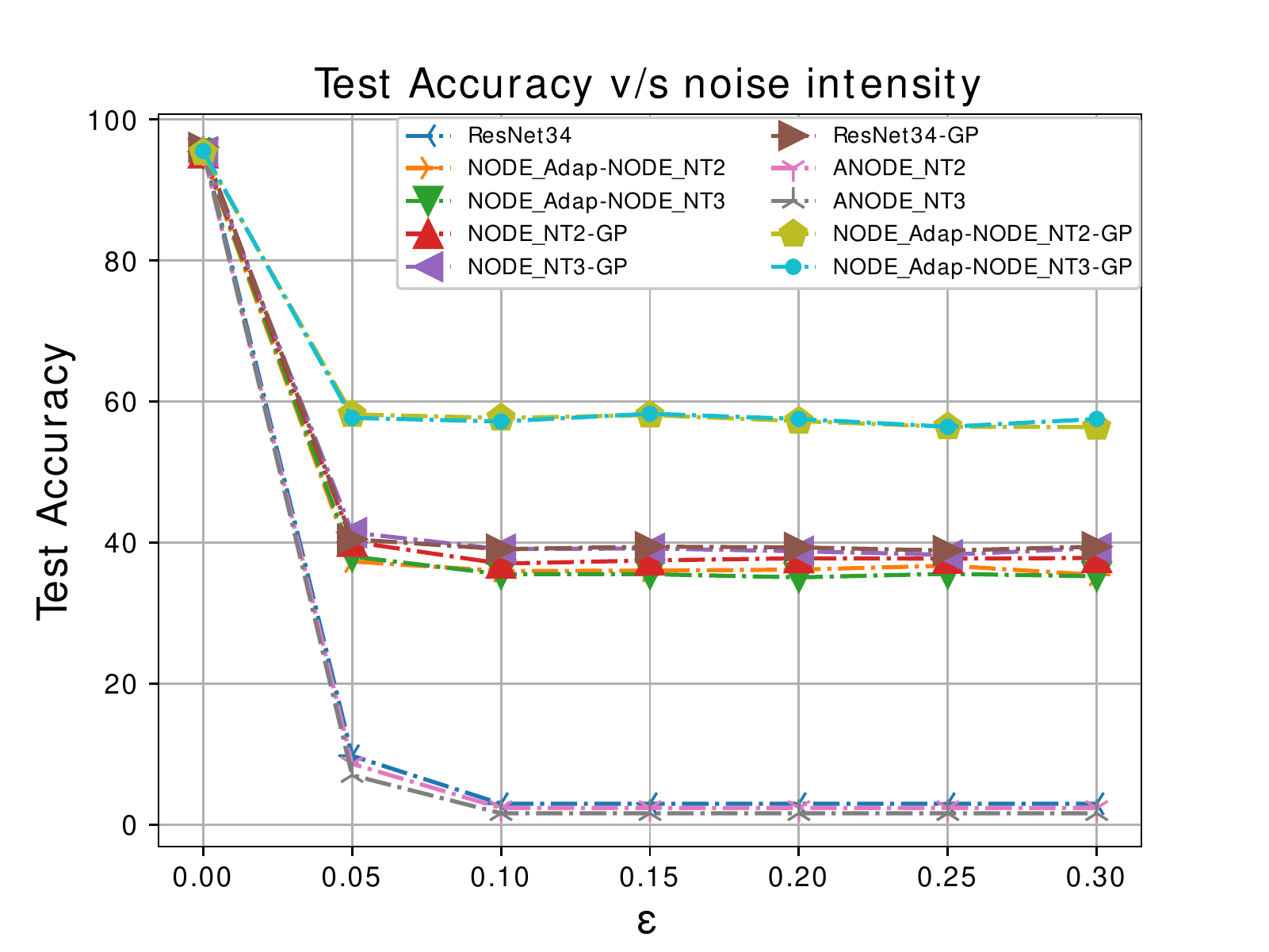}}
\subfigure[FGSM Attack]{\label{fig:b1}\includegraphics[scale=0.38]{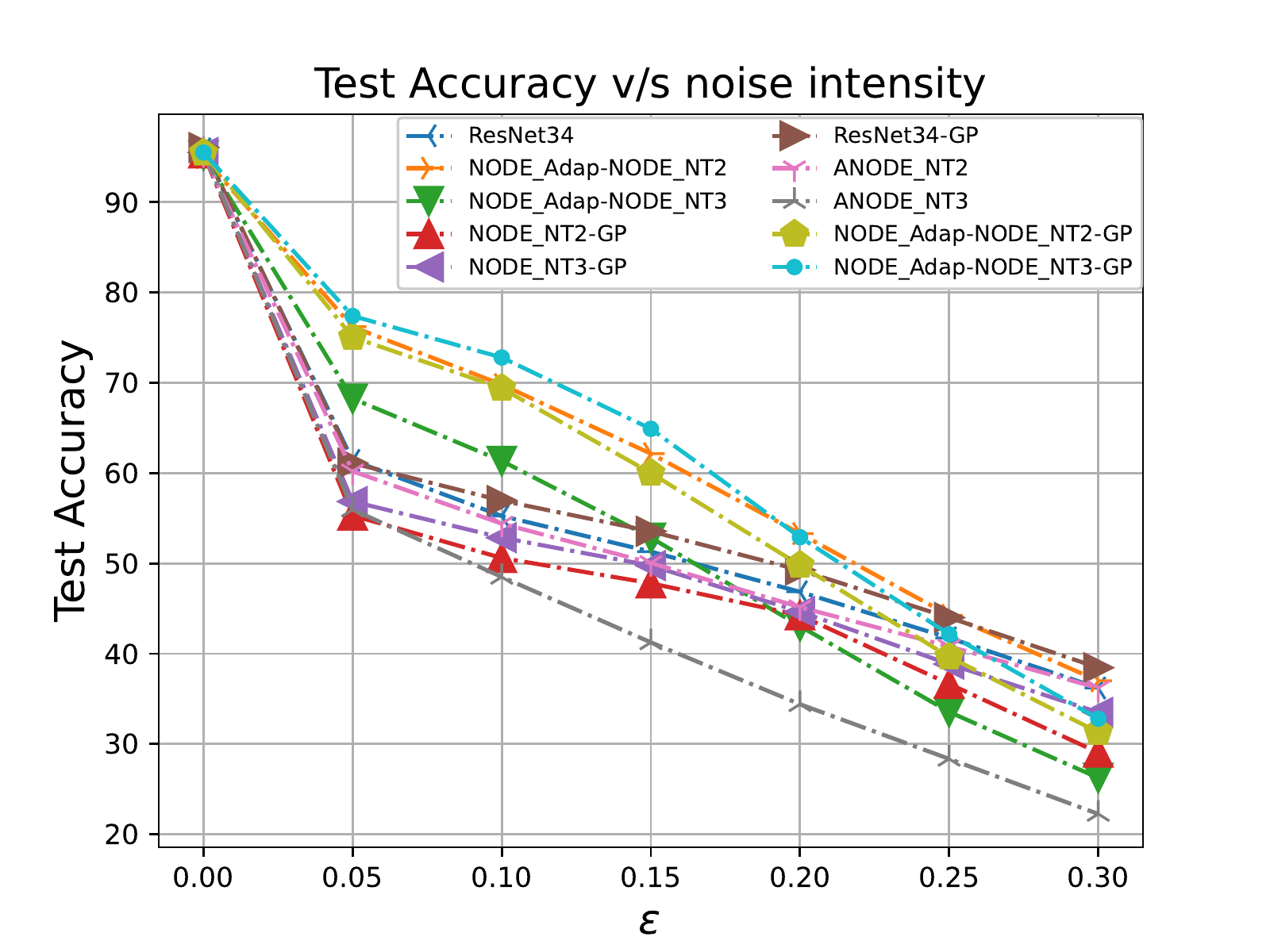}}
\caption{Test accuracy of the models on Cifar10 dataset under (a) PGD attack and (b) FGSM attack }
\label{Adver_cifar10}
\end{figure*}

\section{Experiments}
We conducted experiments to test the ability of the proposed models in terms of modelling  robustness  under adversarial attack and in  capturing uncertainty on out-of-distribution samples~\footnote{Code available at https://github.com/srinivas-quan/NODE-GP.git}. We consider  different variants of the NODE~\footnote{NODE block architecture is similar to the one defined in~\cite{anode}} models and NODE-GP models  considering different numerical techniques in different blocks. For all the NODE models time is varied between 0 and 1.
\begin{itemize}
    \item NODE\_NT2 and NODE\_NT2-GP: This model uses 4 NODE blocks,  with all the NODE blocks using the RK4 numerical solver of two time steps with step size 0.5. The final transformation is done using FCNN in the first and GPs in the second.   
    \item NODE\_NT3 and NODE\_NT3-GP: Here the number of  steps are three instead of two, and step size is 0.33. 
    \item NODE\_Adap-NODE\_NT2 and NODE\_Adap-NODE\_NT2-GP: First block uses adaptive numerical method~\cite{kimura2009dormand}  followed by three NODE blocks which uses RK4 numerical method of time steps two, step size is 0.5. The two methods differ in the final transformation which is FCNN in the former and GPs in the later. 
     \item NODE\_Adap-NODE\_NT3 and NODE\_Adap-NODE\_NT3-GP: Here the number of  steps are three instead of two, and step size is 0.33. 
\end{itemize}
Here, NODE\_NT2 and NODE\_NT3 are the  standard NODE model and are considered as the baseline. Our primary objective of experiments is to show improvements in robustness and uncertainty handling capabilities obtained using  NODE-GP variants. For completeness,  we also compare  against standard  Resnet34 architecture and its variants using GPs in the  final layer (Resnet34-GP)~\cite{SVDKL}.  

\begin{figure*}
\centering     
\subfigure[PGD Attack]{\label{fig:a2}\includegraphics[scale=0.38]{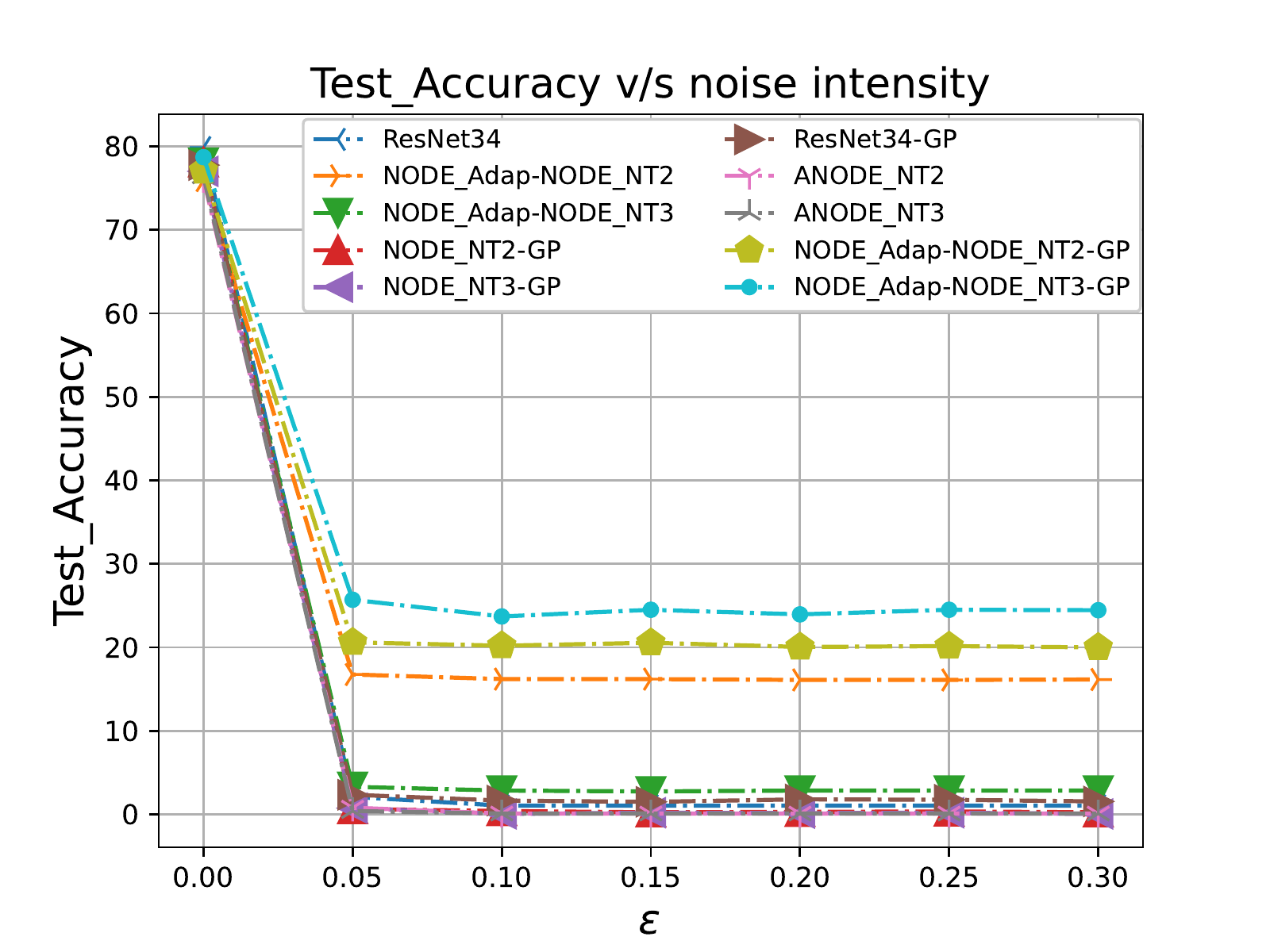}}
\subfigure[FGSM Attack]{\label{fig:b2}\includegraphics[scale=0.38]{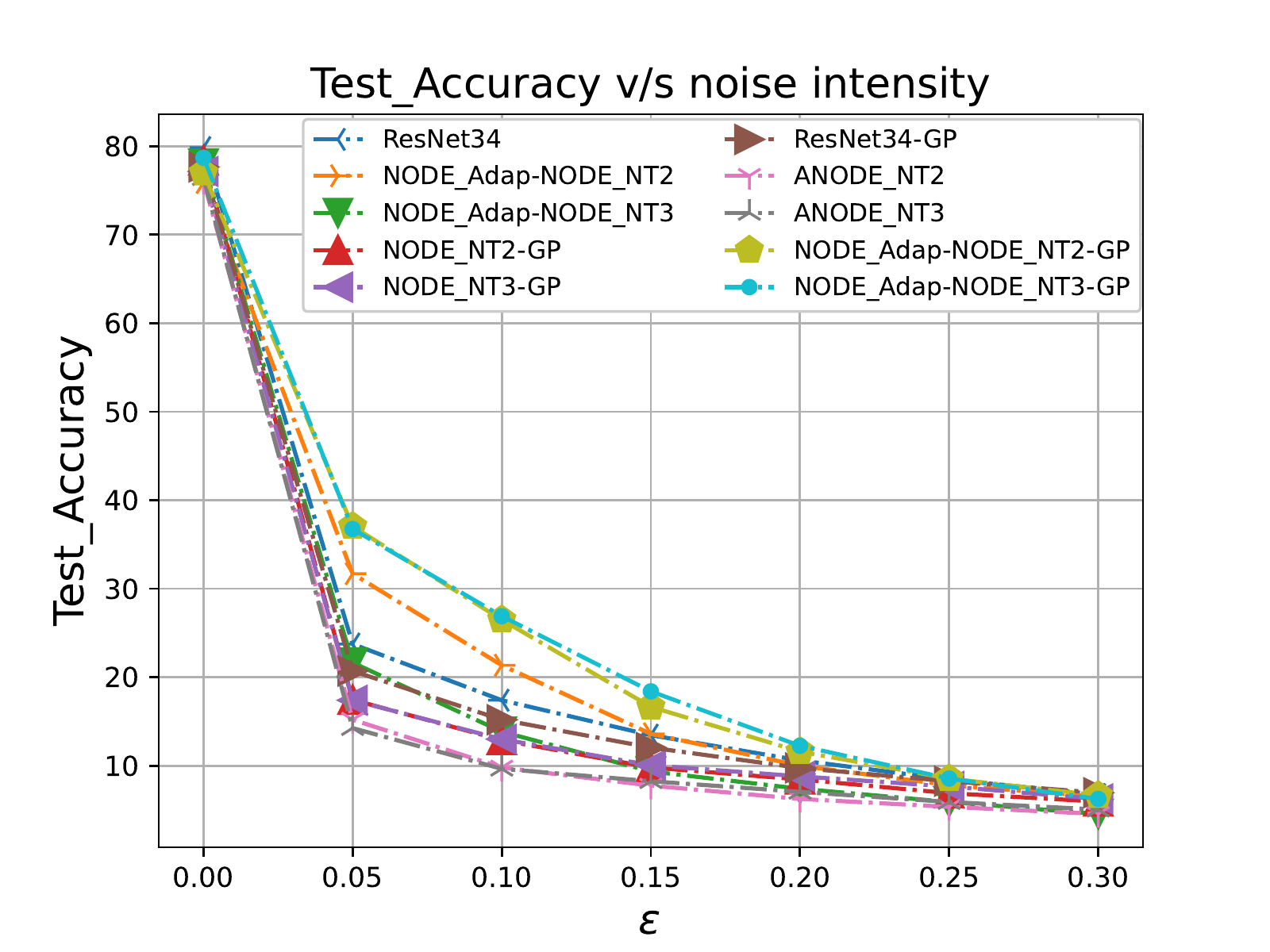}}
\caption{Test accuracy of the models on Cifar100 dataset under (a) PGD attack and (b) FGSM attack }
\label{Adver_cifar100}
\end{figure*}

\subsection{Experimental Setup}
We conduct our experiments on datasets Cifar10~\cite{cifar} with 10 classes and Cifar100~\cite{cifar} with 100 classes. Both the datasets contain 60000 data samples, and we split the samples into 50000  training samples,  5000 validation samples and 5000 samples for testing. Validation dataset is used to choose the model during training.
Every model is trained for 350 epochs with varying learning rate similar to the training of Resnet. For the experiments involving adversarial attack, the accuracy mentioned is for 2000 samples from the test data set. For out of distribution experimental setup, the model trained on Cifar10 is tested with samples of 10 classes from Cifar100 such that the class chosen from cifar100 does not match with any of the classes in cifar10.

\subsection{Adversarial Attacks}
We consider two adversarial attack methods FGSM and PGD discussed in Section \ref{sect:adversarial} to check the robustness of proposed models. 
Figure \ref{Adver_cifar10}(a) and \ref{Adver_cifar10}(b) shows the plot of test accuracy of the models trained on cifar10 with increasing strength $\epsilon$ of the adversarial attacks PGD and FGSM respectively.
The results of PGD and FGSM attacks for Cifar100 is shown in Figure \ref{Adver_cifar100}(a) and \ref{Adver_cifar100}(b)  respectively. 

We can observe that in Figure~\ref{Adver_cifar10} and Figure~\ref{Adver_cifar100}, 
the models NODE\_Adap-NODE\_NT2-GP and NODE\_Adap-NODE\_NT3-GP show better robustness towards attacks compared to all the models. For example, under PGD attack over cifar10 dataset in  Figure~\ref{Adver_cifar10}(a), NODE\_Adap-NODE\_NT3-GP  gives 20\% more accuracy than  NODE\_NT3-GP for $\epsilon \geq0.05$ proving the choice of numerical method matters and gives 22\% more accuracy than NODE\_Adap-NODE\_NT3 for $\epsilon \geq0.05$ proving that inclusion of GP has helped in improving the robustness. In this case,  NODE\_Adap-NODE\_NT2-GP is 52\% more accurate than the baseline standard NODE (NODE\_NT2) and NODE\_Adap-NODE\_NT2 is 30\% more accurate than NODE\_NT2, demonstrating the superior robustness of the proposed NODE models over the standard NODE model. The predictive distribution in GP considers distribution over function values rather than point estimates, helping in reducing the adversarial noise in the image.  The proposed models NODE\_Adap-NODE\_NT2-GP, NODE\_Adap-NODE\_NT3-GP perform better than Resnet variants in almost all values of $\epsilon$ as  observed in Figure~\ref{Adver_cifar10} and Figure~\ref{Adver_cifar100}.  
We can observe from Figure~\ref{Adver_cifar10}(a) and Figure~\ref{Adver_cifar100}(a)  after an $\epsilon$ value we don't see much of a change in accuracy. This is due to the particular nature of the PGD algorithm, where not much change happens to the image after the adversary noise is added  and  projected back to the feasible set around the image. 
\begin{table*}
    \centering
    {\small
    \begin{tabular}{ |p{1.25cm}|p{0.9cm}|p{0.9cm}|p{0.8cm}|p{0.8cm}|p{1.15cm}|p{1.15cm}|p{1.15cm}|p{1.15cm}|p{1.30cm}|p{1.30cm}| }
    \hline 
 \textbf{Dataset} & \textbf{NODE}&\textbf{NODE}&\textbf{Res}&\textbf{Res}&\textbf{NODE}&\textbf{NODE}&\textbf{NODE}&\textbf{NODE}&\textbf{NODE}&\textbf{NODE}\\
 
 \textbf{} & \textbf{\_NT2}&\textbf{\_NT3}&\textbf{net34}&\textbf{net34}&\textbf{\_Adap-}&\textbf{\_Adap-}&\textbf{\_NT2}&\textbf{\_NT3}&\textbf{\_Adap-}&\textbf{\_Adap-}\\
 
 \textbf{} & \textbf{}&\textbf{}&\textbf{}&\textbf{-GP}&\textbf{NODE}&\textbf{NODE}&\textbf{-GP}&\textbf{-GP}&\textbf{NODE}&\textbf{NODE}\\
 
 \textbf{} & \textbf{}&\textbf{}&\textbf{}&\textbf{}&\textbf{\_NT2}&\textbf{\_NT3}&\textbf{}&\textbf{}&\textbf{\_NT2-GP}&\textbf{\_NT3-GP}\\
%
%
  \hline
 \textbf{cifar10}&  95.3 & 95.7 & 95.3 & 95.2 & 95.35 & 95.45 & 95.4 & 95.6 &95.4 & 95.5 \\
  \hline
 \textbf{cifar100}&  76.1 & 76.4 & 78.3 & 76.8 & 76.9 & 76.5 & 75.8 & 77 &76.6 & 76.7 \\
 \hline

 \end{tabular} } 
\caption{Generalization performance of Models in terms of  test accuracies  on Cifar10 and  Cifar100 datasets.}
 \label{tab:accuracy}
\end{table*}

\subsection{Out-of-Distribution Experiments}
\begin{figure}
\centering     
\includegraphics[width=6.cm,height=4.5cm]{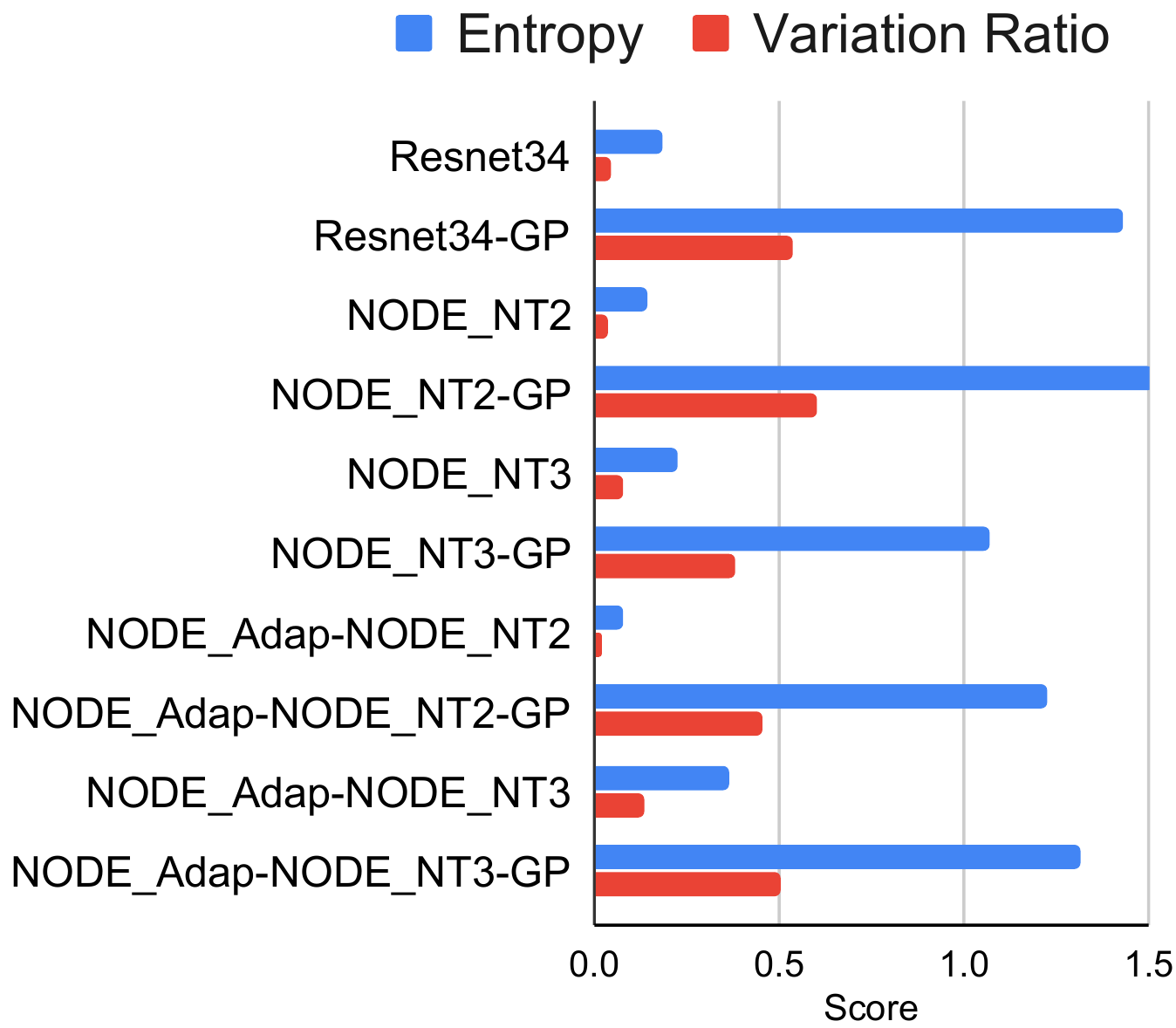}
\caption{Performance of Models on out-of-distribution setting (trained on Cifar10 and tested on Cifar100) using metrics Entropy and Variation Ratio}
\label{out_cifar10}
\end{figure}

We conduct experiments to see the uncertainty modelling capabilities of the proposed models through their performance on out-of-distribution data. 
For the models trained on cifar10, to test the model uncertainty we feed samples from cifar100 dataset.
Figure~\ref{out_cifar10} shows the performance of the models using the metrics Entropy and Variation ratio. For every model, predictive probability is computed on samples chosen from  out-of-distribution data and  an average   Entropy and Variation ratio is computed. Models which use Gaussian processes for final transformation exhibit high Entropy and Variation ratio compared to the models without GP. We find that NODE-GP model, NODE\_NT2-GP exhibited the best uncertainty modelling capabilities in terms of uncertainty and variation ratio. In fact, NODE\_NT2-GP gave entropy score 1.51 and variation ratio score 0.6 as compared to 0.137 and 0.038 by  standard NODE (NODE\_NT2). Thus, the proposed NODE-GP models provide much better uncertainty estimates.

As stated before, the main objective of our work is to improve robustness and uncertainty modeling capabilities in NODE. The experimental results showed that we have indeed achieved our objective. But we also study if this will affect  the generalization performance of the proposed models. The test accuracy  of various  models on Cifar10 and Cifar100 data sets are provided in Table~\ref{tab:accuracy}. We find that the proposed NODE and NODE-GP models achieve an accuracy close to the popular Resnet models. For Cifar10, two of the NODE-GP models gave better performance than standard Resnets. Thus, the proposed models are able to achieve superior robustness and uncertainty modelling capabilities without affecting the generalization performance.  Moreover, NODE and NODE-GP models  have the advantage of less effort in model selection and smaller number of parameters. NODE-GP involves only 13.2 Million parameters compared to  23.4 Million parameters in Resnet34-GP. 

\section{Conclusion}
Neural ODEs provide a continuous time counterpart to Resnets, with advantages in the form of model selection and reduced  number of parameters. However, we found that these models lack robustness and uncertainty handling capabilities  required to help decision making in many real world problems. We propose a novel model combining  Neural ODEs with GPs which will provide the required robustness and uncertainty handling capabilities. Moreover, the model is flexible enough to accommodate different types of NODEs. We also showed that numerical method plays a role in NODE robustness. We consider various types of NODE architectures and numerical solvers to improve the robustness and uncertainty handling capabilities in NODE-GP. The experimental results on Cifar10 and Cifar100 demonstrated the superior performance of the proposed NODE-GP models  against adversarial attacks and uncertainty handling in out-of-distribution data. 


{\small
\bibliographystyle{ieee_fullname}
\bibliography{egbib}
}

\end{document}